\documentclass[runningheads]{llncs}
\usepackage{graphicx}

\usepackage{amsmath,amssymb} 
\usepackage{color}
\usepackage{comment}

\usepackage{subfig}
\usepackage{xspace} 
\usepackage{rotating}

\def\bea{\begin{eqnarray}}
\def\eea{\end{eqnarray}}

\def\mypar#1{{\noindent\bf #1.}}
\def\sect#1{Section~\ref{sec:#1}}

\newcommand{\fig}[1]{Figure~\ref{fig:#1}}
\newcommand{\tab}[1]{Table~\ref{tab:#1}}
\def\Eq#1{Eq.~(\ref{eq:#1})}

\def\ex#1#2{\textrm{I\!E}_{#1}\!\left[#2\right]}     
\def\R{{\rm I\!R}}                                   
\def\Gauss#1#2#3{\mathcal{N}\left(#1 ; #2,#3\right)} 
\def\diag#1{\textrm{diag}\left( #1 \right)}          
\def\iid{i.i.d\onedot}

\makeatletter
\DeclareRobustCommand\onedot{\futurelet\@let@token\@onedot}
\def\@onedot{\ifx\@let@token.\else.\null\fi\xspace}
\def\eg{\emph{e.g}\onedot} 
\def\ie{\emph{i.e}\onedot} 
 
 \def\vs{\emph{vs}\onedot}
 
\def\etal{\emph{et al}\onedot}
\makeatother

\usepackage{array}
\newcolumntype{H}{>{\setbox0=\hbox\bgroup}c<{\egroup}@{}}

\begin{document}
\pagestyle{headings}
\mainmatter

\title{Discrete Point Flow Networks\\
for Efficient Point Cloud Generation}

\titlerunning{Discrete Point Flow Networks for Efficient Point Cloud Generation}
%
\author{Roman Klokov\inst{1}\orcidID{0000-0001-9592-7009} \and
Edmond Boyer\inst{1}\orcidID{0000-0002-1182-3729} 
\and
Jakob Verbeek\inst{2}\orcidID{0000-0003-1419-1816}}
\authorrunning{R. Klokov et al.}
%
\institute{Univ.\ Grenoble Alpes,\ Inria,\ CNRS,\ Grenoble INP,\ LJK,\ 38000 Grenoble,\ France
\email{\{firsname.lastname\}@inria.fr}
\and
Facebook AI Research}

\maketitle
\begin{abstract}
Generative models have proven effective at modeling 3D shapes and their statistical variations. 
In this paper we investigate their application to point clouds, a 3D shape representation widely used in computer vision for which, however, only few generative models have yet been proposed. 
We introduce a latent variable model that builds on normalizing flows with affine coupling layers to generate 3D point clouds of an arbitrary size given a latent shape representation. 
To evaluate its benefits for shape modeling we apply this model for generation,  autoencoding, and single-view shape reconstruction tasks. 
We improve over recent GAN-based models in terms of most metrics that assess generation and autoencoding. 
Compared to recent work based on continuous flows, our model offers a significant speedup in both training and inference times for similar or better performance. 
For single-view shape reconstruction we also obtain results on par with state-of-the-art voxel, point cloud, and mesh-based methods.
\end{abstract}

\begin{figure}
    \centering
    \input{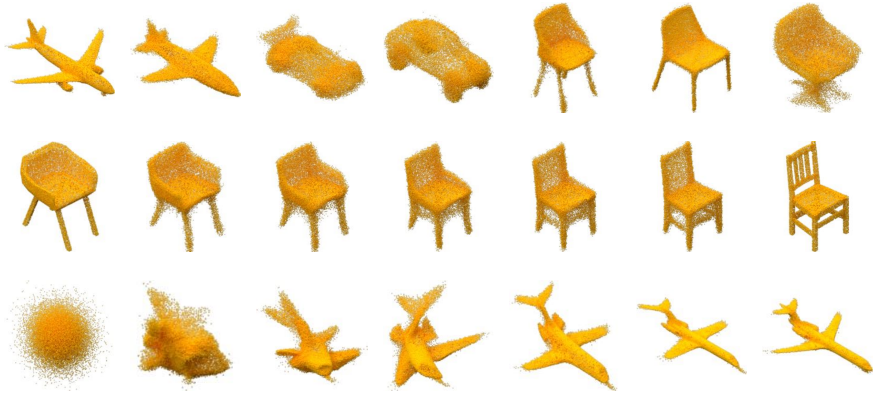}
    \caption{
    Top: Point clouds sampled from DPF-Net for the ShapeNet classes \emph{airplane}, \emph{car}, and \emph{chair}.
    Middle: Latent space interpolation between two point clouds from the test set.
    Bottom: Deformation of points across the flow steps.
    }
    \label{fig:teaser}
\end{figure}
\section{Introduction}
\label{sec:intro}

Generative shape models are used in numerous computer vision applications where they allow to encode 3D shape variations with respect to different attributes, such as shape classes or shape deformations, as well as to infer shapes from partial observations, for instance from a single or a few images. 
Central to shape models is the representation chosen for shapes that can be extrinsic, for example the ubiquitous voxels and octrees, or intrinsic as with meshes and point clouds. 
While extrinsic representations enable relatively straightforward extensions of 2D deep learning techniques to 3D, they suffer from their inherent trade-off between precision and complexity. 
This is why 3D shapes are often represented using intrinsic models, among which point clouds are a natural and versatile solution, serving as a basis for many 3D capturing methods, including most multi-view stereo and range sensing methods, \eg kinect.

Following the success of CNNs for 2D computer vision problems, many deep learning models have been proposed that can handle 3D data. 
This includes works on voxel grids \cite{brock16arxiv,choy16eccv,girdhar16eccv,graham18cvpr,klokov19bmvc,maturana15iros,wu16nips,wu15cvpr}, octrees \cite{riegler17cvpr,tatarchenko17iccv}, meshes \cite{henderson18bmvc,monti17cvpr,verma18cvpr,wang18eccv}, point clouds \cite{fan17cvpr,insafutdinov18nips,klokov17iccv,mandikal18bmvc,qi17cvpr,qi17nips,wang19ijcai}, and implicit functions \cite{michalkiewicz19iccv,park19cvpr}. 
While they provide effective tools to build predictive models of 3D shapes, \eg from a single image, we investigate in this paper the less extensively explored and more generic problem of probabilistic generative 3D shape modeling.

Significant advances have been made in generative modeling of natural images using deep neural networks with convolutional architectures. 
Consequently, they can easily be adapted to generative shape models which are based on extrinsic representations, using regular 3D convolutional layers \cite{wu16nips,brock16arxiv,klokov19bmvc}. 
On the other hand, their extensions to intrinsic representations, such as point clouds and meshes, are less obvious and, to the best of our knowledge, so far, only Yang \etal~\cite{yang19iccv} have studied generative models from which arbitrary size point clouds can be sampled without any conditioning information.

We explore a hierarchical latent variable model that treats the points as exchangeable variables, which allows us to model and sample point clouds of arbitrary size. 
Within this framework, each point cloud is considered as a sample from a shape-specific distribution over the 3D surface of the object, and these distributions are embedded in a latent space. 
To sample a point cloud, first, a vector is sampled in the latent shape space, and then, any desired number of 3D points can be sampled \iid conditioned on the latent shape representation. 
Our model shares the high-level structure with PointFlow~\cite{yang19iccv}, but differs in the underlying network architectures, reducing the training and sampling time by more than an order of magnitude. 
In particular, our model builds on discrete normalizing flows with affine coupling layers \cite{dinh17iclr} rather than continuous flows, and FiLM conditioning layers \cite{perez18aaai} to construct a flexible density on 3D points given the latent shape representation. 
In \fig{teaser} we illustrate diverse point clouds sampled from our class-specific models, interpolation between point clouds in the learned latent shape space, and the sequential process by which the discrete normalizing flow warps the points to obtain the final shape.

We evaluate generative and autoencoding capabilities of our model, as well as its use for single-view shape reconstruction. 
We obtain similar or better performance compared to GAN-based models in terms of metrics that assess generation and autoencoding. 
Compared to recent work based on continuous flows, our model offers a significant speedup, for similar or better performance. 
For single view reconstruction our model performs on par with state-of-the-art methods, yet allows to reconstruct with arbitrarily large point clouds. 
Moreover, we analyze various design choices regarding data splitting and normalization.

\section{Related work}
\label{sec:related}

\mypar{Generative Models}
Deep neural networks have sparked significant progress in generative modeling. 
The most widely adopted models are variational autoencoders (VAEs) \cite{kingma14iclr,rezende14icml}, generative adversarial networks (GANs) \cite{goodfellow14nips,karras18iclr}, and normalizing flows \cite{deco95nips,dinh17iclr}. 
All three approaches share the basic principle of defining a latent variable $z$ with a simple prior, \eg unit Gaussian, and construct a complex conditional $p(x|z)$ on data $x$ by means of deep neural networks. 
Maximum likelihood training of the resulting marginal $p(x)$ is generally intractable due to the non-linearities. 
To train the model, VAEs rely on an amortized inference network that produces a variational posterior $q(z|x)$. 
GANs, on the other hand, use a discriminator network to distinguish training examples and model samples, and use it as a signal to train the generative model. 
Alternatively to previous approaches, normalizing flow models rely on invertible neural network architectures to avoid the intractability of the marginalization altogether. 
In this case, the likelihood can be computed exactly by the means of the change of variable formula, and latent variables can be inferred deterministically. 
A variety of different normalizing flows has recently been proposed, see \eg \cite{behrmann19icml,chen18nips,dinh17iclr,grathwohl19iclr,kingma18nips,kingma16nips,rezende15icml}. 
See~\cite{kobyzev19arxiv,papamakarios19arxiv} for recent comprehensive reviews on normalizing flows.

Affine coupling layers \cite{dinh17iclr} allow for a computation of the inverse in a closed form that is as efficient as the function itself. 
Within this approach, the activations $A^{\ell}$ in layer $\ell$ are partitioned in two groups, $A^{\ell}_1$ and $A^{\ell}_2$. 
The first group is unchanged, and used to update the other group by scaling and translation, \ie $A^{\ell+1}_2 = A^{\ell}_2 \odot s(A^{\ell}_1) + t(A^{\ell}_1)$, where $\odot$ denotes element-wise multiplication, and $s(\cdot)$ and $t(\cdot)$ can be arbitrary (non-invertible) neural networks. 
The inverse is trivially obtained by subtraction and division, since $A^{\ell+1}_1  =  A^{\ell}_1$. 
Many coupling layers with changing variable partitioning can be stacked to construct a complex invertible flow. 
Training and sampling the model require to compute the flow reverse directions, and since affine coupling layers are equally efficient in both directions, it means that both processes are fast. 
This is in contrast to some other normalizing flows, such as invertible ResNets \cite{behrmann19icml,chen18nips}, or planar and radial flows \cite{rezende15icml}, for which the inverse flow does not have a closed form.

Neural ordinary differential equations were recently proposed as a generalization of deep residual networks (ResNets, \cite{he16cvpr,he16eccv}) in the limit of infinite depth~\cite{chen18nips}. 
Chen \etal \cite{chen18nips} demonstrated that Neural ODEs can be used to define normalizing flow, which are referred to as ``continuous normalizing flows''.

Several conditional flow-based models have recently been proposed for vision tasks. 
In \cite{lu20aaai} flows conditioned on an input image are used for image segmentation, inpainting, denoising, and depth refinement. Their model is trained directly via maximum likelihood estimation, as their model does not include a global latent variable. 
Similarly, C-Flow~\cite{pumarola20cvpr} does not involve a global latent variable, and rather than treating point clouds as sets, it sorts the points to a regular pixel grid, and applies 2D normalizing flows for single image point cloud reconstruction. 
A conditional VAE model, where flow is used to define a flexible distribution on the latent variable given the conditioning data was introduced by \cite{bhattacharyya19nipsws}. 
This is similar in structure to our model for single view reconstruction. Their experiments, however, concern the prediction of point trajectories in 2D for hand-written digits, and traffic participants such as pedestrians and cars. 
The generative image model of~\cite{lucas19nips} is related to ours, as a VAE model with a flow-based decoder. 
The application contexts, RGB images \vs point clouds, and resulting architectures are, however, quite different.

\mypar{Point Cloud Generating Networks}
Deep learning models for point cloud processing have received significant attention in recent years, both in recognition \cite{qi17cvpr,klokov17iccv,qi17nips,zaheer17nips,groueix18cvpr,su18cvpr} and generative \cite{li18arxiv,achlioptas18icml,yang19iccv} settings. 
The PointNet architecture of Qi \etal \cite{qi17cvpr} was the first to propose a deep network for recognition of point clouds. 
The points are first processed in an identical and independent manner by an MLP, and global max-pooling is used to aggregate the per-point information. 
KD-Net \cite{klokov17iccv} and PointNet++ \cite{qi17nips} add a notion of spatial proximity to the architecture, replacing global max-pooling with local aggregation. 
While these models can interpret point clouds, they cannot generate them.

Early point cloud generating networks \cite{achlioptas18icml,fan17cvpr} produce point clouds with a fixed number of points $n$, by using a network with $n\times 3$ outputs. 
AtlasNet~\cite{groueix18cvpr} mitigates this limitation by using a set of $k$ square 2D patches, and deforming each of these non-linearly by using $k$ patch-specific MLPs that takes as input 2D patch coordinates as well as a global shape representation. 
The shape vector is obtained from a point cloud encoder network (for autoencoding), or from a CNN trained for single-view image reconstruction. 
The point cloud GAN (PC-GAN, \cite{li18arxiv}) is related, but uses a single generator that that takes a global shape vector as input together with (arbitrarily many) samples from a unit Gaussian. 
Similarly to \cite{fan17cvpr}, AtlasNet is a conditional model, that generates point clouds \emph{given} another point cloud or an image. In contrast, PC-GAN includes a second generator that models a distribution on the latent shape space, so it can generate point clouds in an unconditional manner.

The high-level hierarchical latent variable structure of PointFlow \cite{yang19iccv} is similar to PC-GAN. 
Rather than using adversarial training, however, they train the model using a VAE-like approach in which an inference network produces an approximate posterior on the latent shape representation. 
Moreover, they use continuous normalizing flows~\cite{chen18nips} to define a prior on the shape space, and a conditional distribution on 3D points given a latent shape representation. 
Our work is based on the same high-level VAE-like structure as PointFlow, but differs in the design of the  network components. 
Most importantly,  we make use of efficient ``discrete'' affine coupling layers, avoiding the use of computationally expensive ODE solvers for training and generation needed for the ``continuous'' flows, resulting in a significant speed-up to train and sample from the model. 
We describe our ``Discrete Point Flow Networks'' in the next section.

\section{Discrete Point Flow Networks}
\label{sec:method}

In this section we first present the high-level hierarchical latent variable model, followed by a more detailed description of the model components in  \sect{method:details}.

\subsection{Hierarchical Latent Variable Model for Point Cloud Generation}
\label{sec:method:hierarchical}

Our goal is to define a generative model over point clouds of variable size that represent 3D shapes. 
The defining characteristics of point clouds are that the number of points may vary from one cloud to another, and that there is no inherent ordering among the points.

Let $X=\{x_1,\dots, x_n\}$ be a point cloud with $x_i\in\R^d$, where $d=3$ for point clouds for 3D shapes. 
The dimension $d$ may be larger in some cases, \eg $d=6$ when modeling 3D points equipped with surface normals. 
An exchangeable distribution is one that is invariant to permutations of the data, \ie
\bea
p(x_1,\dots, x_n) = p(x_{\pi_1},\dots, x_{\pi_n}),
\eea
where $\pi$ is a permutation of the integers $1,\dots,n$. 
Note that independence implies exchangeability, but the reverse does not hold.

De Finetti's representation theorem states that any exchangeable distribution can be written as a factored distribution, conditioned on a latent variable:
\bea
p(X) = \int_z p_{\psi}(z) \prod_{x\in X} p_{\theta}(x|z) \textrm{d}z.
\label{eq:definetti}
\eea
In the case of 3D point cloud modeling, the latent variable $z$ can be thought of as an element in an abstract shape space, sampled from a prior $p_{\psi}(z)$. 
This construction allows for point clouds of different cardinality, since conditioned on the shape representation $z$, the elements of the point cloud are sampled \iid. 
Given this general framework, also adopted in \cite{li18arxiv,yang19iccv}, the challenge is to:
\begin{enumerate}
    \item 
    Design a flexible model so that the conditional distribution $p_{\theta}(x|z)$ concentrates around the surface of the object represented by $z$.
    \item
    Mitigate the intractability of the integral in \Eq{definetti} during training when using, \eg, deep neural networks to construct $p_{\theta}(x|z)$.
\end{enumerate}
Before we consider the design of $p_{\theta}(x|z)$ and $p_{\psi}(z)$ in \sect{method:details}, we describe how to deal with the integral in \Eq{definetti} using the VAE framework~\cite{kingma14iclr}.

We efficiently approximate the intractable posterior $p(z|X)$ with an amortized inference network $q_\phi(z|X)$. 
The approximate posterior allows us to define a variational bound on the likelihood in \Eq{definetti} as using Jensen's inequality~\cite{bishop06patrec}:
\bea
\ln p(X) &\geq & 
\sum_{x\in X} \ex{q_\phi(z|X)}{\ln p_\theta(x|z)} - \mathcal{D}_\textrm{KL}(q_\phi(z|X)||p_\psi(z))\equiv - \mathcal{F}.
\label{eq:fe}
\eea
The first term aims to reconstruct the points $x\in X$ using shape representations sampled from $q_\phi(z|X)$, whereas the second term ensures that the approximate posterior cannot arbitrarily deviate from the prior. 
Following \cite{kingma14iclr,rezende14icml} we use Monte Carlo sampling and the reparametrization trick to jointly minimize the loss $\mathcal{F}$ over $\theta$, $\psi$ and $\phi$ using stochastic gradient descent. 
The distributions $q_\phi(z|X)$, $p_\theta(x|z)$, and $p_\psi(z)$ and the underlying network architectures that make up the loss are detailed in the following section.

\subsection{Design of Model Components}
\label{sec:method:details}

\mypar{Shape-conditional Point Distribution}
The density on points for a given latent shape, $p_{\theta}(x|z)$, needs to be flexible enough to concentrate its support around the surface of the 3D shape. 
To this end we construct a conditional form of normalizing flows based on affine coupling layers \cite{dinh17iclr}.

Let $y\in\R^3$ denote a latent variable for each 3D point $x$, with a Gaussian conditional distribution given by  $p_{\theta}(y|z)=\Gauss{y}{\nu_\theta(z)}{\diag{\omega_\theta(z)}}$, where $\nu_\theta(z)$ and $\omega_\theta(z)$ are non-linear functions of $z$. 
In the affine coupling layer, we partition the coordinates of $y$ in two groups, $y^c$ and $y^u$, and update $y^u$ by affine transformation conditioned on $y^c$ and the latent shape representation $z$, \ie $x^u = y^u \odot s_\theta(y^c,z) + t_\theta(y^c,z)$, while leaving the conditioning coordinates unchanged, \ie $x^c = y^c$. 
To achieve the desired expressivity, we stack many affine coupling layers, cycling through the six possible partitionings of the three coordinates. 
Each coupling layer in the resulting flow $f_\theta(x;z)$ is conditioned on the latent shape representation $z$ by the means of the FiLM conditioning mechanism \cite{perez18aaai} in the scaling and translation functions.

\begin{figure}[t]
    \centering
    \resizebox{.9\textwidth}{!}{
    \includegraphics[width=\textwidth]{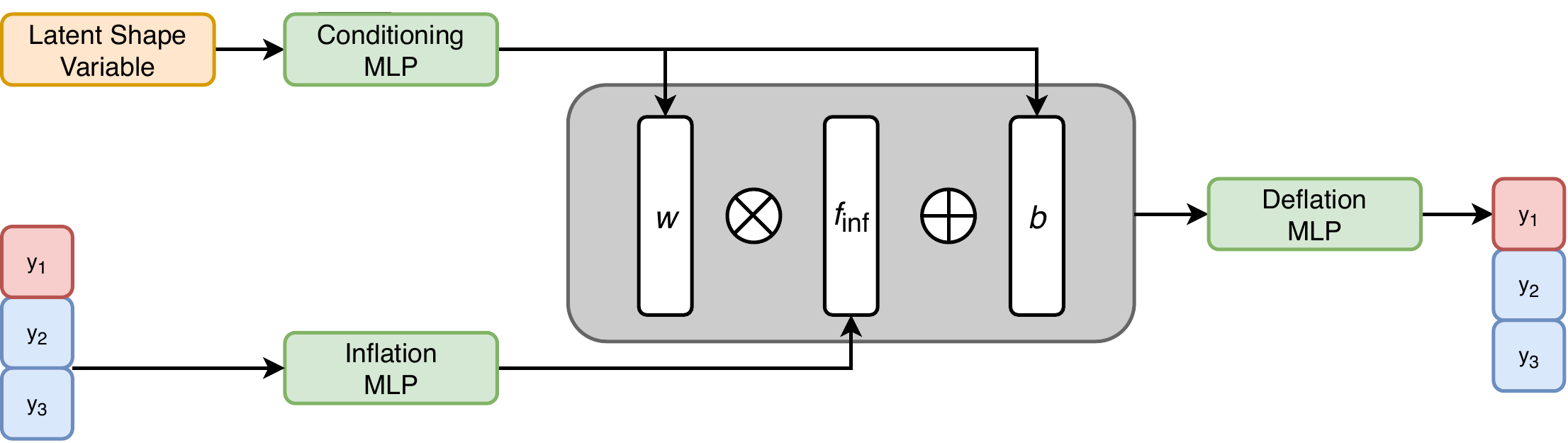}
    }
    \caption{Architecture of our conditional affine coupling layer applied to a single 3D point, with red dimension of the point being updated given the blue ones.}
    \label{fig:coupling}
\end{figure}

In practice, the scaling and translation functions are implemented by MLPs, which inflate the dimensionality of $y^c$ to $D_{\textrm{inf}}$, and then deflate it to the dimensionality of $y^u$. 
Simultaneously, a separate MLP takes the latent variable $z$ and outputs conditioning coefficients of size $D_{\textrm{inf}}$, with which we multiply and shift the inflated hidden units in the scaling and translation functions. 
In \fig{coupling} we provide an overview of the architecture of our conditional coupling layers.

Using $f_\theta(x;z)$ to denote the invertible flow network that maps $x$ to $y$, the change of variable formula allows to write the density of 3D points $x$ given $z$ as:
\bea
p_{\theta}(x|z) = \Gauss{f_\theta(x;z)}{\nu_\theta(z)}{\diag{\omega_\theta(z)}}
\; \left| 
\textrm{det}\left(\frac{\partial f_\theta(x;z)}{\partial x^\top}\right)
\right|,
\eea
which enters into the loss defined in \Eq{fe}.

\mypar{Amortized Inference Network}
The amortized inference network $q_\phi(z|X)$ takes a point cloud and produces a distribution on the latent shape representation. 
We use a permutation invariant design based on the PointNet architecture for shape classification \cite{qi17cvpr}. 
As an output, the model produces the mean and diagonal covariance matrix of a Gaussian on $z\in\R^D$, \ie $q_{\phi}(z|X)=\Gauss{z}{\mu_\phi(X)}{\diag{\sigma_\phi(X)}}$.

\mypar{Latent Shape Prior}
Rather than using a unit Gaussian prior in the latent space, as is common in deep generative models, we learn a more expressive prior $p_\psi(z)$ by means of another normalizing flow $g_\psi(z)$ based on affine coupling layers, similar to \cite{chen17iclr,yang19iccv}. 
In our experiments it reduces the KL divergence in \Eq{fe} by adapting the prior to fit the marginal posterior $\sum_X q_\phi(z|X)$, rather than forcing the inference network to induce a unit Gaussian marginal posterior, resulting in improved generative performance. 
Using this construction, we obtain the KL divergence as:
\bea
\mathcal{D}_\textrm{KL}(q_\phi(z|X)||p_\psi(z)) 
 = 
\ex{q_\phi(z|X)}{\ln p_\psi(z)} -\mathcal{H}(q_\phi(z|X)) \\
 =   
\ex{q_\phi(z|X)}{\ln \Gauss{g_\psi(z)}{\eta}{\diag{\kappa}} + \ln \left|\textrm{det}\left(\frac{\partial g_\psi(z)}{\partial z^\top}\right)\right|} -1^\top\ln\sigma_\phi(X),
\eea
where we use Monte Carlo sampling to approximate the expectation.

\begin{figure}[t]
    \centering
        \resizebox{1.0\textwidth}{!}{
    \includegraphics[width=\textwidth]{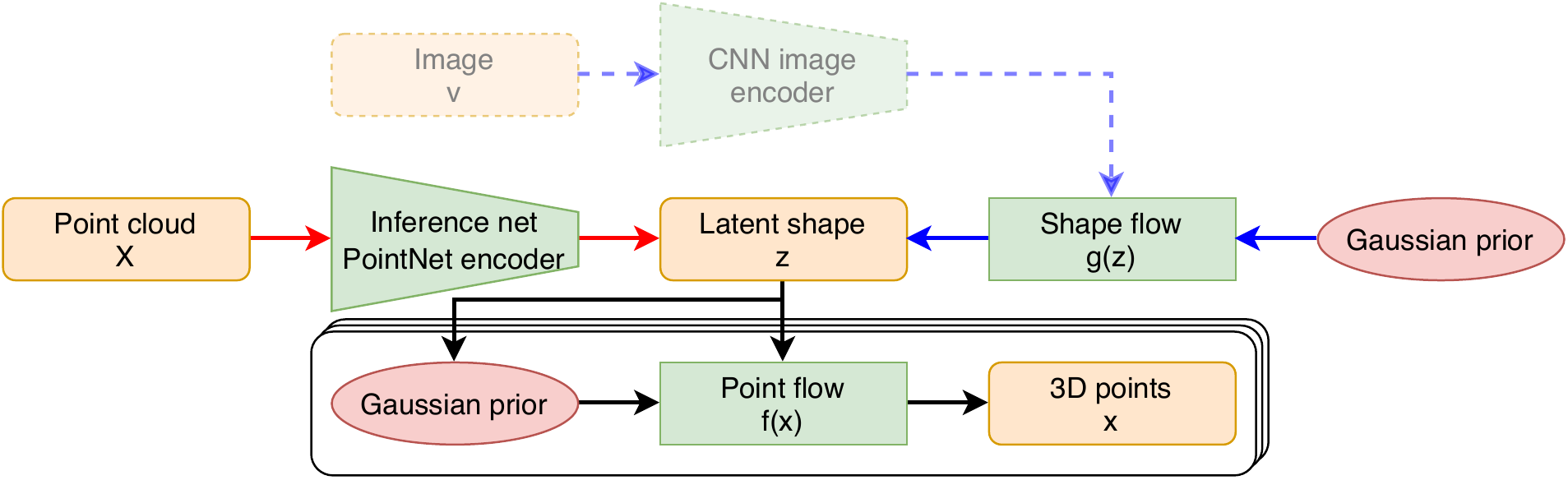}
    }
    \caption{Overview of DPF-Net: arrows indicate data flow to sample new point clouds (blue) and point cloud autoencoding (red), black arrows are used in both processes. During training flow modules are traversed in the reverse direction. For single-view reconstruction the shape prior is conditioned on the image (dashed).
    }
    \label{fig:model}
\end{figure}

\mypar{Single-View Reconstruction Architecture}
For single-view reconstruction, we follow \cite{klokov19bmvc} and define the model as:
\bea
p(X|v) = \int_z p_\psi(z|v) \prod_{x\in X} p_\theta(x|z) \textrm{d} z,
\eea
where we replaced the latent shape prior $p_\psi(z)$ with an image conditioned one, $p_\psi(z|v)$. 
In this case, the latent shape flow $g_\psi$ does not deform a parametric Gaussian, but rather a Gaussian whose mean and variance are computed from an image $v$ by a CNN encoder. 
We train the model by optimizing a variational bound similar to \Eq{fe}, and using the PointNet inference network to obtain an approximate posterior. 
\fig{model} provides an overview of the data flow between the model components for training and point cloud generation.

\section{Experiments}
\label{sec:exp}

\mypar{Datasets}
In order to provide a comparison with prior academic studies on shape generation, we perform experiments on subsets of ShapeNet~\cite{chang15arxiv} dataset. 
For autoencoding we use the ShapeNetCore.v2, containing roughly 55k meshes from 55 classes. 
In the generative setting, following \cite{yang19iccv}, we use single class subsets (\emph{airplanes}, \emph{cars}, and \emph{chairs}) from the same dataset. 
For the single-view reconstruction task we used a subset of 13 major classes of ShapeNetCore.v1 from Choy \etal \cite{choy16eccv}, which comes with rendered $137\times 137$ images from 24 randomized viewpoints per shape. 
We substitute the voxel grids provided by Choy \etal with the original meshes to sample point clouds for training and evaluation.

\mypar{Data Split}
For autoencoding we use a random split of the data, distributed across train, validation, and test sets in a 70/10/20 proportion per class. 
In the generative setting, we use single class subsets from the same random split. 
By using a random data split per class we intentionally deviate from the official ShapeNet data split, used in \cite{yang19iccv}, which splits into significantly different data subsets per class. 
For example, in case of airplanes, training and validation sets mostly contain regular passenger aircraft, while the test set is populated with fighter jets and spaceships. 
While such a split could be useful in the context of autoencoding to assess out-of-distribution generalization, a significant mismatch between training and test set is undesirable for evaluation of generative models which are supposed to fit the training distribution. 
For single-view reconstruction we use the train/test split from \cite{choy16eccv}.

\mypar{Normalization}
The original meshes in the ShapeNet are not normalized for position and scale which negatively affects the reconstruction quality. 
We therefore, additionally use a normalized version of the dataset, where we preprocess each mesh separately so that the sampled point clouds are (approximately) zero mean, and tightly fit in a unit diameter sphere. 
For generative experiments we use models trained and evaluated on normalized data. 
In case of the autoencoding, we report results for two separate DPF-Nets trained either on non-normalized or normalized data, where in the latter case we rescale point clouds to original scales before evaluation for comparability with the rest of the models. 
While using non-normalized data, similarly to \cite{yang19iccv}, we perform global normalization across all shapes by translation to the aggregate center of all the training shapes, but do not rescale to unit global variance. 
For single-view reconstruction we compare models trained on normalized data, but evaluate them in the unit radius sphere scale for comparability with related work.

Point clouds are uniformly sampled from the meshes by sampling polygons with a probability proportional to their area, and then uniformly sampling a point per each selected polygon. 
Unlike previous works using precomputed point clouds, we perform this procedure on the fly, thus obtaining a different random point cloud each time we process a 3D shape. 
During both training and evaluation we sample two point clouds for each shape: one is used as input to the inference network, the other for optimization or evaluation of the decoder. 
We use $2,048$, $2,048$, and $2,500$ points for training and quantitative evaluation for generative, autoencoding, and single-view reconstruction tasks accordingly.

\mypar{Evaluation Metrics}
We follow the standard protocol \cite{achlioptas18icml,fan17cvpr,yang19iccv} and use Chamfer distance (CD) and earth mover's distance (EMD) to assess point cloud reconstructions. 
To measure the generative properties we follow \cite{achlioptas18icml,yang19iccv}, and use metrics to compare  equally sized \emph{sets} of generated and reference point clouds:
\begin{itemize}
    \item The Jensen-Shannon divergence (JSD) compares the marginal distributions obtained by taking the union of all generated (or reference) point clouds, and quantizing the distributions to a voxel grid.
    \item The Minimum matching distance (MMD) computes the average distance of reference point clouds to their nearest (in CD/EMD) generated point cloud.
    \item Coverage (COV) is the fraction of reference point clouds  matched by  minimum CD/EMD distance by at least one  generated point cloud.
    \item 1-nearest neighbour  accuracy (1-NNA) classifies generated and reference point clouds as belonging to either of these two sets using a leave-one-out 1-nearest neighbor classifier (using CD/EMD). Ideal accuracy is 50\%.
\end{itemize}
For single-view reconstruction we additionaly report F1-score. 
For more detailed descriptions of these metrics we refer to \cite{achlioptas18icml,yang19iccv}.

\subsection{Experimental Setup}
\mypar{Inference Network}
We use a PointNet encoder \cite{qi17cvpr} with the number of features progressing over layers as $3-64-128-256-512$, followed by a max-pooling across points, and two fully-connected layers of sizes $512$ and $D$, where $D$ is the size of the latent space. 
We use $D = 512$ in the autoencoding and single-view reconstruction experiments, and $D = 128$ for generative modeling.

\mypar{Latent Shape Prior}
We use $14$ affine coupling layers to construct the latent space prior. 
In the coupling layers, we alternate between two orthogonal partitioning schemes: odd and even dimensions are split in different groups; the first $D/2$ dimension go in one group, and the remaining in the other. 
For single-view reconstruction we use ResNet18 image encoder, similarly to \cite{groueix18cvpr,wang18eccv,wang19ijcai}.

\mypar{Point Decoder}
The point decoder $p(x|z)$ starts with a three-dimensional Gaussian with mean and variances computed from $z$ by an MLP with two hidden layers. 
This Gaussian is transformed by $63$ of our conditioned affine coupling layers, each consisting of (i) two fully-connected layers that map $z$ to the FiLM conditioning coefficients, each of size $64$, (ii) two fully-connected layers that inflate input dimensionality to $64$ hidden units (at which point the FiLM conditioning is performed), (iii) a final fully-connected layer that deflate the dimensionality to compute the scaling and translation functions.

\mypar{Baselines}
We retrained AtlasNet \cite{groueix18cvpr}, l-GAN-CD/EMD \cite{achlioptas18icml}, PointFlow \cite{yang19iccv}, and DCG \cite{wang19ijcai} with our split of the ShapeNet dataset, using the implementation provided by the authors and our data processing pipeline. 
For improved comparability we also modified the point cloud encoders in all models to match each other (except for l-GANs, since it significantly worsened their results). 
To match other approaches, we used $2,048$ points per cloud for AtlasNet in the autoencoding task and consequently set the number of learned primitives to $16$.

\mypar{Oracles}
For all the tasks we also provide an ``oracle'' to assess the best possible performance values. 
For autoencoding and single-view reconstruction, the oracle samples a second point cloud from the ground truth mesh, rather than generating a point cloud. 
For generative modeling, the oracle uses the point clouds from the training set, instead of sampling point clouds from the model.

\noindent Our code is publicly available at \url{https://github.com/Regenerator/dpf-nets}.

\subsection{Generative Modeling Evaluation}
\label{sec:generative}

\mypar{Efficiency Comparison with PointFlow}
We compare our DPF-Networks in terms of computational efficiency and memory footprint to PointFlow \cite{yang19iccv}. 
We train both models for point cloud generation, and report the number of parameters and total training time. 
To compute the training memory footprint, training time, and generation time per sample  (point cloud), we divided total GPU memory occupied during training, batch iteration time, and batch generation time, respectively, by the batch size.

\begin{table}[b]
  \begin{center}
    \caption{Efficiency comparison for DPF-Nets and PointFlow generative models.}
    \label{tab:efficiency}
    \resizebox{.9\textwidth}{!}{
\setlength{\tabcolsep}{2pt}
\begin{tabular}{|l|c|c|c|c|c|}
\hline
Model   & Nr.\  params., & Mem.\ footprint,    & tr. time, & total tr. time,   & gen. time, \\
        & $10^6$        & Mb/sample         & ms/sample & days              & ms/sample \\
\hline
\hline
PointFlow \cite{yang19iccv}  & 1.63            & 470                     &  500         &  80           &  150          \\
DPF-Net (Ours) & 3.76            &  370                     & 16                       & 1.1                       & 4                         \\
\hline
\end{tabular}
}
  \end{center}
\end{table}

Both models were run on a single TITAN RTX GPU. 
We estimate the total training time for PointFlow after the initial $100$ epochs of training, which took 2 days, and assume that the full training procedure requires $4,000$ epochs, as reported by the authors of PointFlow. 
We observed that the training procedure slowed down over the course of training, because ODE-solver gradually increases the number of iterations to meet the required tolerance. 
Thus, all timings in \tab{efficiency} for PointFlow should be understood as lower bounds. 

From the results in \tab{efficiency} we see that even though DPF-Networks have more parameters, the associated training memory footprint is lower and, our model is approximately $30$ times faster both in training and inference iterations, and can be trained in a single day.

\mypar{Quantitative Results}
We compare to l-GANs and PointFlow models and report oracle performance as a reference. 
Given the prohibitive computation cost of complete PointFlow training, we provide results obtained after training for four days, which is four times the full training time of DPF-Net in the same setting. 
In order to account for random sampling every model is evaluated using ten different sets of generated objects, each of the size of the test set. 
Thus, for each metric we report mean values over ten runs. 
In addition to the best values, in \tab{generation} we also write in bold results that are within two standard deviations of the best result.

\begin{table}[t]
  \begin{center}
    \caption{
    Generative modeling results. 
    Oracle results are underlined when the are not the best. 
    JSD and MMD-EMD are multiplied by $10^2$, MMD-CD by $10^4$.
    }
    \label{tab:generation}
    \resizebox{.9\textwidth}{!}{
\setlength{\tabcolsep}{5pt}
\begin{tabular}{|l|l|c|c|c|c|c|c|c|}
\hline
    &   & JSD$\downarrow$ & \multicolumn{2}{c|}{MMD$\downarrow$} & \multicolumn{2}{c|}{COV$\uparrow$, \%} & \multicolumn{2}{c|}{1-NNA$\downarrow$, \%} \\
Category & Model &                                        & CD               & EMD              & CD         & EMD        & CD         & EMD        \\
\hline
\hline
         & l-GAN-CD \cite{achlioptas18icml}  & 2.76       & {\bf 5.69}       & 5.16             & 39.5       & 17.1       & 72.9       & 92.1       \\
         & l-GAN-EMD \cite{achlioptas18icml} & 1.77       & 6.05             & {\bf 4.15}       & 39.7       & 40.4       & 75.7       & 73.0       \\
Airplane & PointFlow \cite{yang19iccv}       & 1.42       & 6.05             & 4.32             & {\bf 44.7} & {\bf 48.4} & {\bf 70.9} & {\bf 68.4} \\
         & DPF-Nets (Ours)                   & {\bf 0.94} & 6.07             & 4.26             & {\bf 46.8} & {\bf 48.4} & {\bf 70.6} & {\bf 67.0} \\
         & Oracle                            & 0.50       & \underline{5.97} & 3.98             & 51.4       & 52.7       & 49.8       & 48.2       \\
\hline
\hline
         & l-GAN-CD \cite{achlioptas18icml}  & 2.65       & {\bf 8.83}       & 5.36             & 41.3       & 15.9       & {\bf 62.6} & 92.7       \\
         & l-GAN-EMD \cite{achlioptas18icml} & 1.31       & 9.00             & {\bf 4.40}       & 38.3       & 32.9       & 65.2       & {\bf 63.2} \\
Car      & PointFlow \cite{yang19iccv}       & 0.59       & 9.53             & 4.71             & {\bf 42.3} & 35.8       & 70.1       & 74.2       \\
         & DPF-Nets (Ours)                   & {\bf 0.45} & 9.59             & 4.61             & {\bf 43.4} & {\bf 45.7} & 70.3       & {\bf 64.3} \\
         & Oracle                            & 0.37       & \underline{9.24} & \underline{4.56} & 52.8       & 52.7       & 50.9       & 50.5       \\
\hline
\hline
         & l-GAN-CD \cite{achlioptas18icml}  & 3.65       & {\bf 16.66}      & 7.91             & 42.3       & 17.1       & 68.5       & 96.5       \\
         & l-GAN-EMD \cite{achlioptas18icml} & 1.27       & {\bf 16.78}      & {\bf 5.75}       & 44.3       & 43.8       & 66.6       & 67.8       \\
Chair    & PointFlow \cite{yang19iccv}       & 1.51       & 17.15            & 6.20             & 43.3       & 46.5       & 67.0       & 70.4       \\
         & DPF-Nets (Ours)                   & {\bf 1.01} & 17.08            & 6.14             & {\bf 46.9} & {\bf 48.5} & {\bf 63.5} & {\bf 64.8} \\
         & Oracle                            & 0.49       & 16.39            & 5.71             & 52.8       & 53.4       & 49.7       & 49.6       \\
\hline
\end{tabular}
}
  \end{center}
\end{table}

Overall, DPF-Networks yield the best results in terms of JSD, COV-CD/EMD and 1-NNA-CD/EMD, except for the 1-NNA-CD for \emph{car}. 
This confirms that our DPF-Network is capable of generating more realistic and diverse sets of point clouds, for random samples from our model see \fig{teaser}.

L-GAN-CD experiences mode collapses, and generates objects with good CD values, but with very poor coverage and 1-NNA in terms of the EMD metric. 
PointFlow shows performance similar to ours, except for JSD, while being significantly slower in both training and sampling. 
In contrast to the evaluations performed in \cite{yang19iccv}, based on the official split, in our experiments the oracle obtains the best  performances for all metrics, except for MMD (see underlined results). 
We believe that this highlights the fact the MMD metric does not favor diversity in the generated point clouds, but instead favors point clouds with low CD/EMD distances to all the reference shapes. 
If the generated point clouds contain a subset of high quality modes from the test subset, the metric can yield good results, even better than the oracle. 
DPF-Nets and PointFlow yield qualitatively similar point cloud samples, we provide a comparison of samples in the supplementary material.

\subsection{Autoencoding Evaluation}
\label{sec:autoencoding}

\begin{table}[t]
  \begin{center}
    \caption{Autoencoding results. $\dagger$ - results from \cite{yang19iccv} on the official split, $*$ - results for equal training time as DPF-Net on the random split.
    }
    \label{tab:autoencoding}
    \resizebox{.9\textwidth}{!}{
\setlength{\tabcolsep}{32pt}
\begin{tabular}{|l|c|c|}
\hline
Metric                                  & CD $\times10^4$   & EMD $\times10^2$      \\
\hline
\hline
l-GAN-CD \cite{achlioptas18icml}        & 7.07              & 7.70                  \\
l-GAN-EMD \cite{achlioptas18icml}       & 9.18              & 5.30                  \\
AtlasNet \cite{groueix18cvpr}           & {\bf 5.66}        & 5.81                  \\
PointFlow$^\dagger$ \cite{yang19iccv}   & 7.54              & 5.18                  \\
PointFlow$^*$                           & 10.22             & 6.58                  \\
DPF-Net, orig.                          & 6.85              & 5.06                  \\
DPF-Net, norm.                          & 6.17              & {\bf 4.37}            \\
Oracle                                  & 3.10              & 3.13                  \\
\hline
\end{tabular}
}
  \end{center}
\end{table}

\begin{figure}[b]
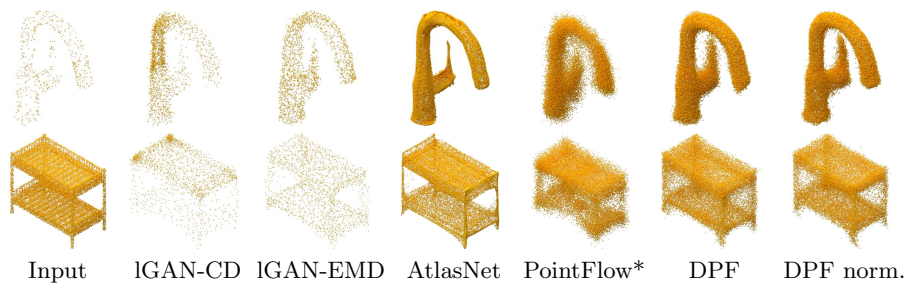

    \centering
    
\def\myfig#1#2{{\includegraphics[width=0.125\textwidth]{#1_ae_#2_sqr.jpeg}
}}
{
\begin{tabular}{ccccccc}
\small
\myfig{sparse}{gt_4800} &
\myfig{sparse}{lgancd_4800} &
\myfig{sparse}{lganemd_4800} &
\myfig{sparse}{an_4800} &
\myfig{sparse}{pf_4800} &
\myfig{sparse}{dpfn_4800} &
\myfig{sparse}{dpfnn_4800} \\
\myfig{dense}{gt_1110}&
\myfig{dense}{lgancd_1110} &
\myfig{dense}{lganemd_1110} &
\myfig{dense}{an_1110} &
\myfig{dense}{pf_1110} &
\myfig{dense}{dpfn_1110} &
\myfig{dense}{dpfnn_1110} \\
Input &
lGAN-CD & 
lGAN-EMD & 
AtlasNet & 
PointFlow* & 
DPF & 
DPF norm. 
\end{tabular}
}
    \caption{Qualitative comparison of the models from \tab{autoencoding} for the autoencoding task with sparse (top) and dense (bottom) inputs.}
    \label{fig:ae_samples}
\end{figure}

We compare DPF-networks with other models in terms of autoencoding performance in \tab{autoencoding}. 
Similarly to generative experiments, we restricted the training time of PointFlow, this time, to match the training time of our approach which was approximately a week. 
Among models trained on non-normalized data, DPF-Net (orig.) achieve the best results in the EMD metric and second best in the CD metric, outperformed only by the non-generative AtlasNet which is trained by optimization of the CD metric. 
The DPF-Net outperforms both l-GANs which were specifically optimized for the CD/EMD metrics, while being trained by optimization of the likelihood lower bound. 
Importantly DPF-Nets outperform PointFlow in both metrics under the same and extended computational budget.

When our model is trained on normalized data, results significantly improve, achieving state-of-the-art among generative models for both metrics. 
This underlines the importance of proper data normalization for shape modeling.

In \fig{ae_samples} we qualitatively compare our autoencoding results to l-GANs, AtlasNet, and PointFlow. 
All approaches can work with arbitrary size inputs, but only AtlasNet, PointFlow, and DPF-Nets can reconstruct with arbitrary density. 
In this comparison we use $512$ and $32,768$ points as sparse and dense inputs, while reconstructing fixed $2,048$ points for l-GANs and $32,768$ point for AtlasNet, PointFlow and DPF-Nets. 
Models with better CD values (l-GAN-CD, AtlasNet) tend to concentrate points in some regions of reconstructed shapes, while models with better EMD values (l-GAN-EMD, DPF-Nets) distribute points more evenly. 
While AtlasNet achieves best CD, its reconstructions contain sharp plane-like artifacts. 
Our DPF-Nets produce overall smoother reconstructions, but, on the other hand, suffer from more noise.

\begin{figure}[t]
    \begin{center}
    
\def\myfig#1{{\includegraphics[width=0.105\textwidth]{svr_#1_sqr.jpeg}
}}

{
\begin{tabular}{cccccccc}
\small
\myfig{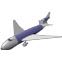} &
\myfig{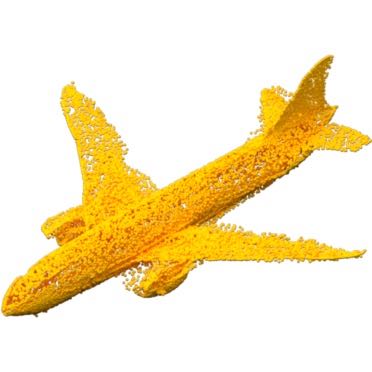} &
\myfig{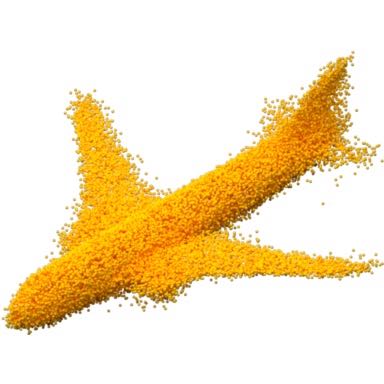} &
\myfig{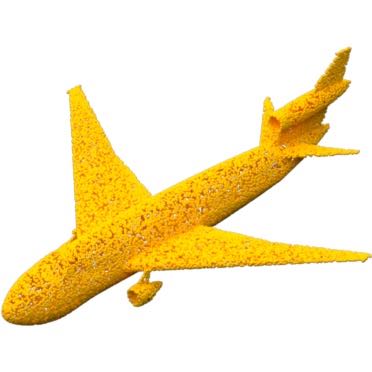} &
\myfig{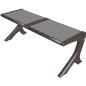} &
\myfig{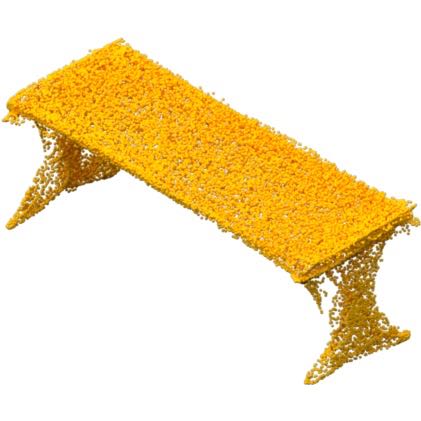} &
\myfig{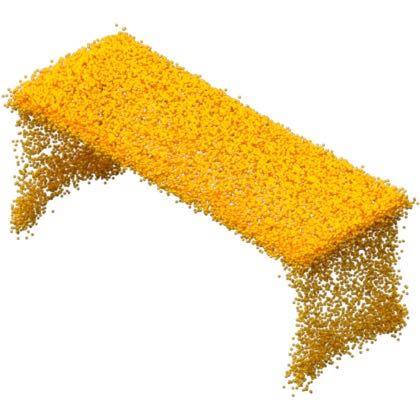} &
\myfig{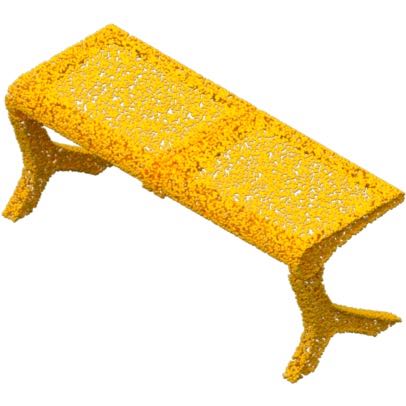} \\
\myfig{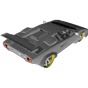} &
\myfig{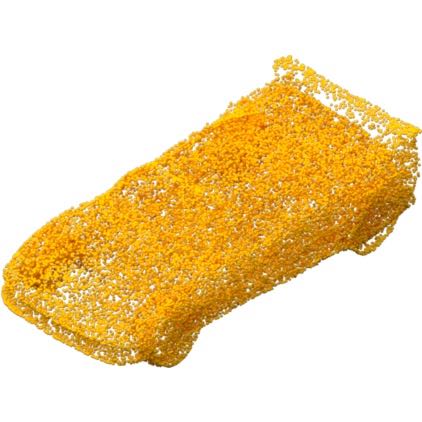} &
\myfig{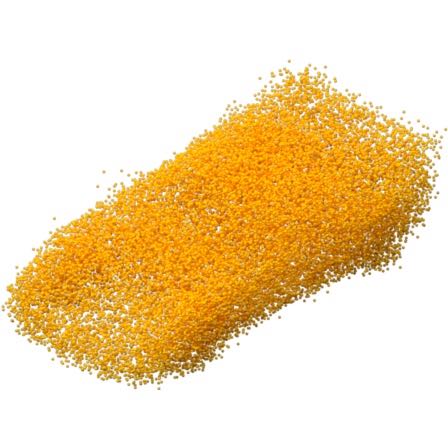} &
\myfig{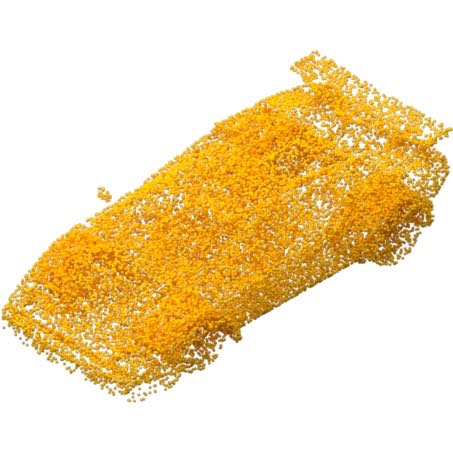} &
\myfig{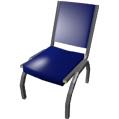} &
\myfig{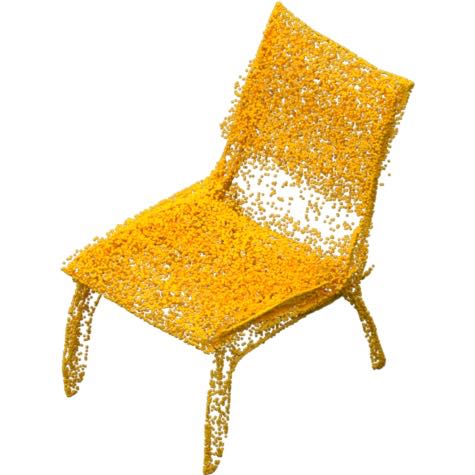} &
\myfig{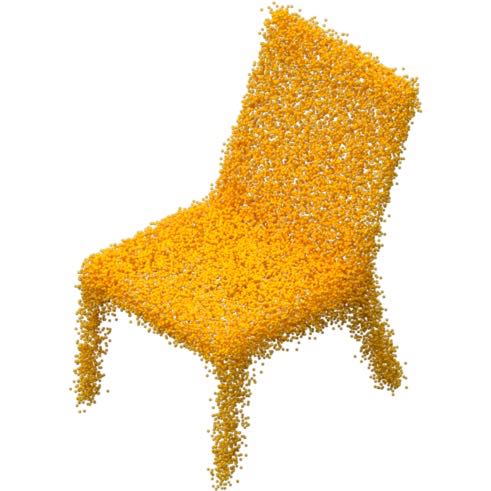} &
\myfig{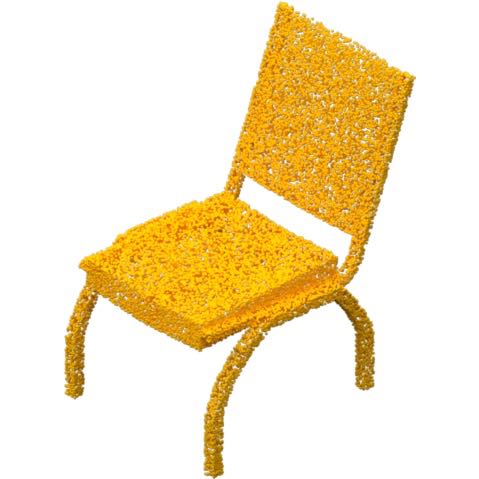} \\
\myfig{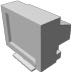} &
\myfig{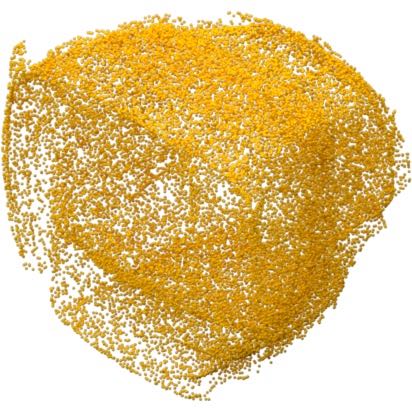} &
\myfig{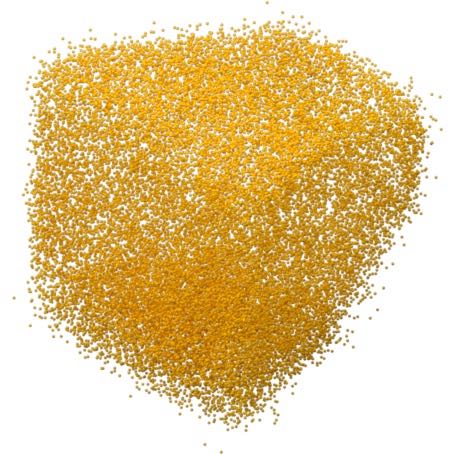} &
\myfig{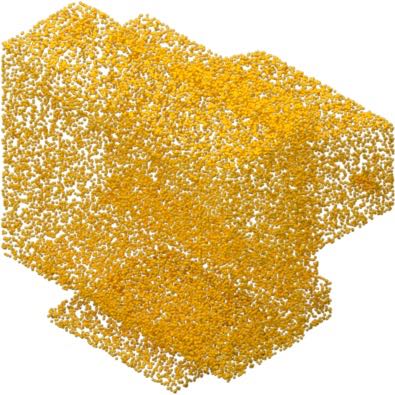} &
\myfig{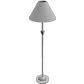} &
\myfig{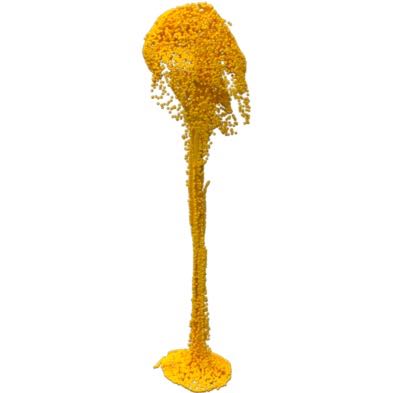} &
\myfig{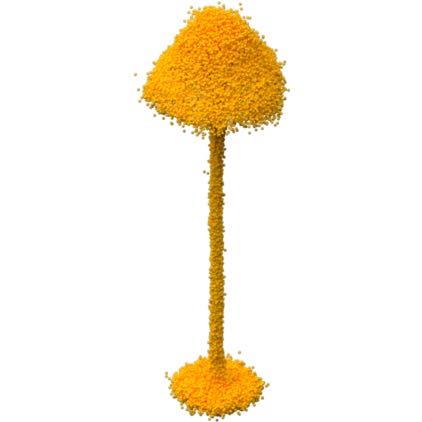} &
\myfig{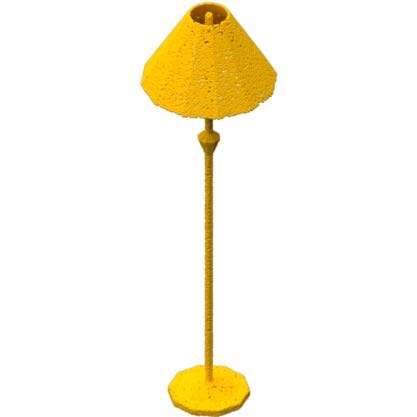} \\
\myfig{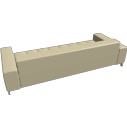} &
\myfig{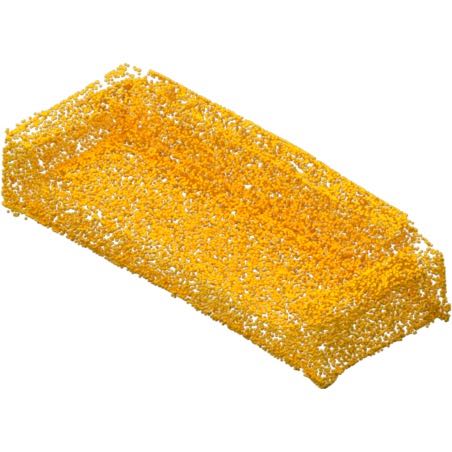} &
\myfig{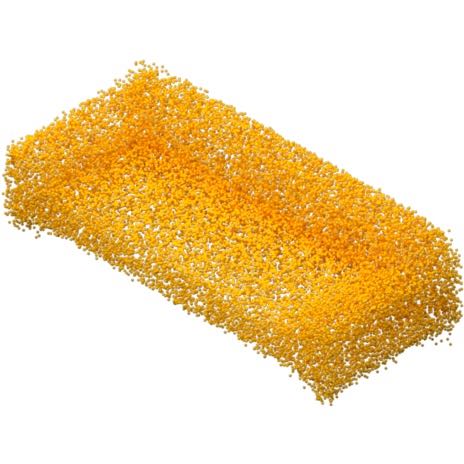} &
\myfig{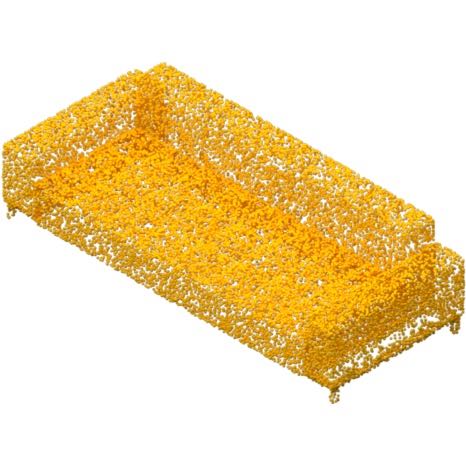} &
\myfig{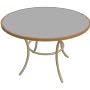} &
\myfig{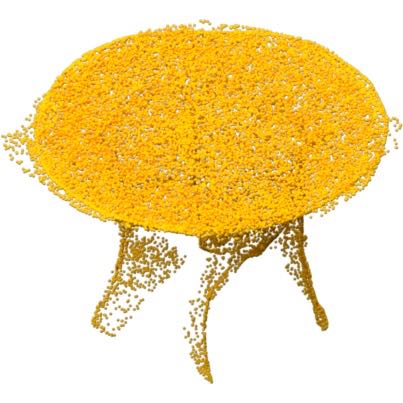} &
\myfig{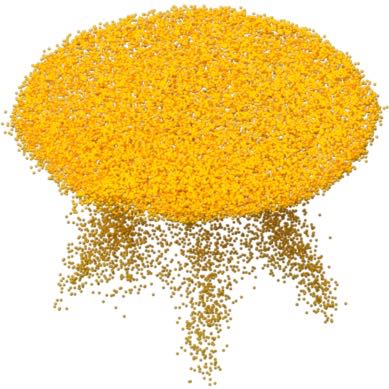} &
\myfig{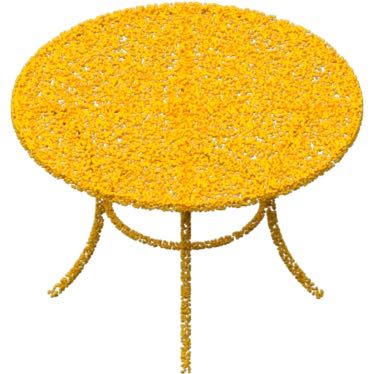} \\
Input &
AtlasNet & 
DPF-Net & 
gr.\ truth &
Input &
AtlasNet & 
DPF-Net & 
gr.\ truth 
\end{tabular}
}
    \end{center}
    \caption{Qualitative comparison for single-view reconstruction task. 
    }
    \label{fig:SVR}
\end{figure}

\subsection{Single-View Reconstruction}
In this section we test DPF-Nets on the inference of 3D point clouds from single images. 
The architecture used for this specific task is depicted in Figure~\ref{fig:model} and detailed in \sect{method:details}. We compare our results to recent state-of-the-art methods in the field. 
This includes: the voxel-based PRN~\cite{klokov19bmvc}, point cloud-based approaches of AtlasNet~\cite{groueix18cvpr} and DCG~\cite{wang19ijcai}, and the mesh-based Pixel2Mesh~\cite{wang18eccv}.

Although convenient, in general, comparison to voxel-based approaches should be taken with a grain of salt, since it is biased. 
To compute the proposed metrics either ground truth or reconstructed voxelized shapes are fed to the marching cubes algorithm to obtain final meshes which are used to sample point clouds. 
Resulting ground truth meshes in that case are crude approximations of the original meshes, used in the evaluation of the point cloud and mesh-based approaches. 
Moreover, there are cases both in voxelized data and reconstructions, when the marching cubes algorithm fails to output meshes.

The results in \tab{svr} show that DPF-Net clearly outperforms earlier works in terms of the EMD metric. 
It also achieves best results in terms of the F1-score among point cloud and mesh-based models. 
In terms of CD, similarly to autoencoding it is outperformed only by AtlasNet with a small margin. 
This validates the ability of normalizing flows to capture complex distributions in 3D and to model shape surfaces.

Qualitative single-view reconstruction results can be found in \fig{SVR}. 
Note that a single reconstruction model has been trained across all 13 classes for both AtlasNet and DPF-Net. 
Similarly to the autoencoding task, compared to AtlasNet our approach produces more evenly distributed point clouds without sharp dense clusters, but introduces more noise.

\begin{table}[t]
  \begin{center}
    \caption{
    Single-view reconstruction results. 
    $^{\dagger}$: results taken from \cite{wang18eccv}.
    }
    \label{tab:svr}
    \resizebox{.9\textwidth}{!}{
\setlength{\tabcolsep}{16pt}
\begin{tabular}{|l|c|c|c|}
\hline
Model                                       & CD$\downarrow$, $\times 10^3$     & EMD$\downarrow$, $\times 10^2$  & F1$\uparrow, \tau = 0.001, \%$  \\
\hline
\hline
PRN \cite{klokov19bmvc}                     & 7.56                              & 11.00                           & {\bf 53.1}                  \\
AtlasNet \cite{groueix18cvpr}               & {\bf 5.34}                        & 12.54                           & 52.2                        \\
DCG \cite{wang19ijcai}                      & 6.35                              & 18.94                           & 45.7                        \\
Pixel2Mesh$^{\dagger}$ \cite{wang18eccv}    & 5.91                              & 13.80                           & -                           \\
DPF-Nets                                    & 5.51                              & {\bf 10.95}                     & 52.4                        \\
Oracle                                      & 1.10                              & 5.70                            & 84.0                        \\
\hline
\end{tabular}
}
  \end{center}
\end{table}

\section{Conclusion}
\label{sec:conclusion}

We presented DPF-Networks, a generative model for point clouds of arbitrary size. 
DPF-Nets are based on a latent variable model and use normalizing flows with affine coupling layers to construct a flexible, yet tractable, shape conditional density on 3D points, and an expressive latent shape space prior. 
They are trained akin to VAEs, using a permutation invariant point cloud encoder as approximate posterior distribution over the latent shape space.

The evaluation on the ShapeNet dataset demonstrates that DPF-nets improve generative performance metrics over previous work in most metrics and classes. 
Compared to a recent related work based on continuous normalizing flows, our model is between one and two orders of magnitude faster to train and sample from. 
Applied to single view reconstruction, DPF-Nets outperform state-of-the-art methods, hence showing promising capabilities in 3D shape modeling.

\smallskip
\mypar{Acknowledgements}
Work done while Jakob Verbeek was at INRIA.

\appendix
\section{Training Details}
\label{sec:training}
All the models were trained with AMSGrad optimizer \cite{reddi18iclr} with decoupled weight decay regularization \cite{loshchilov19iclr} with step-like schedule for the learning rate. All the hyperparameters for all the experiments can be found on the paper webpage \url{https://github.com/Regenerator/dpf-nets}.

\section{Detailed Generative Modeling Performance}
\label{sec:gend}
In \tab{gen_ext} we present the same evaluation as in Table 2 of the main paper, but include the standard deviations across the ten sets of point clouds sampled from each model.

\section{Qualitative Results for Autoencoding}
\label{sec:qual}
In \fig{sparse_ae_samples} and \fig{dense_ae_samples} we show reconstruction results for l-GANs, AtlasNet, PointFlow, and ours DPF-Net from sparse and dense input point clouds. 
All the models used 512 and 32,768 points as a sparse or dense input accordingly. 
Note, that AtlasNet, PointFlow and DPF-Net are able to reconstruct arbitrary densely (32,768 points here), for l-GANs output size is fixed to 2,048.

\section{Qualitative Results for Generation}
\label{sec:add_smp}
In figures \ref{fig:samples_air}---\ref{fig:samples_cha} we provide additional samples from the models trained for the Airplane, Car, and Chair classes. 
For each point cloud we sample $32,768$ points.

\section{Additional interpolations}
\label{sec:add_ints}
\begin{figure}[h!]
    \centering
    \input{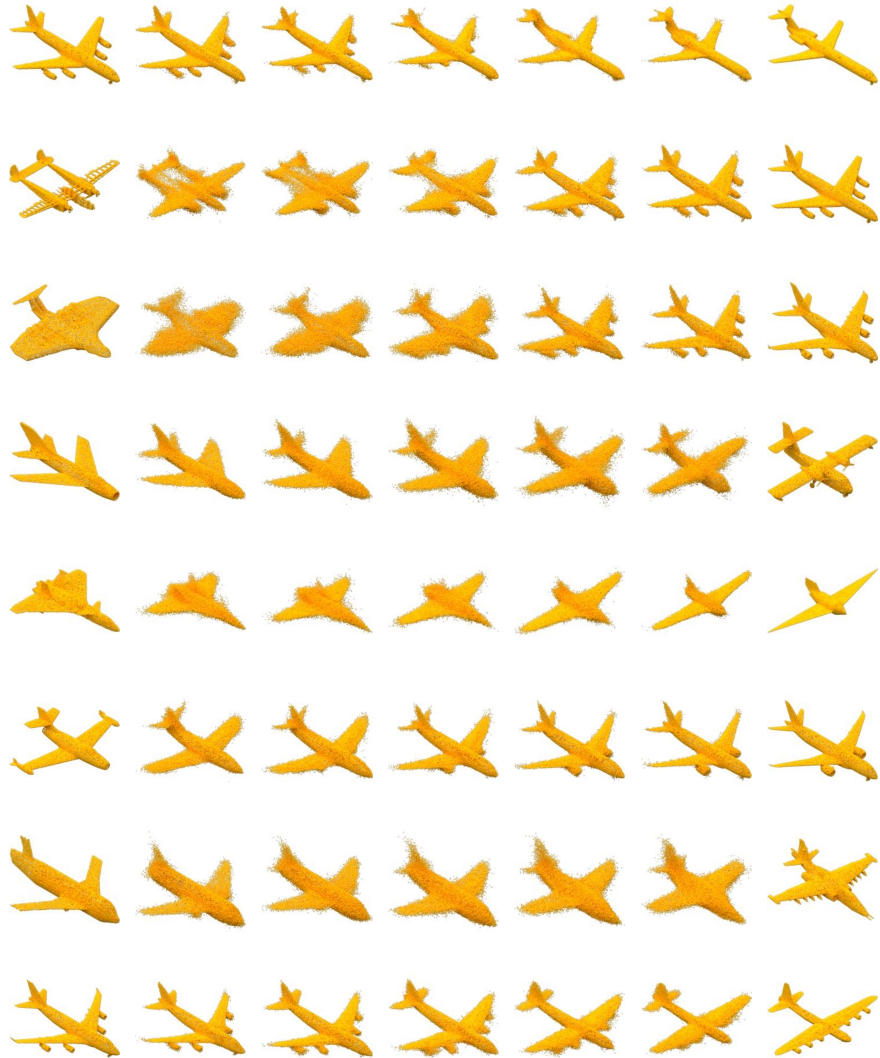}
    \caption{Interpolations between Airplane data samples.}
    \label{fig:int_air}
\end{figure}

In figures \ref{fig:int_air}---\ref{fig:int_cha} we provide latent space interpolations from the models trained for the Airplane, Car, and Chair classes. 
We sample two shapes from the test data (left- and right-most in the figures), and interpolate between corresponding latent variable samples to obtain a path in the shape space. 
For each latent shape on the path we then sample a point cloud of size $32,768$.

\section{Additional flow illustrations}
\label{sec:add_flows}
\begin{figure}[h!]
    \centering
    \input{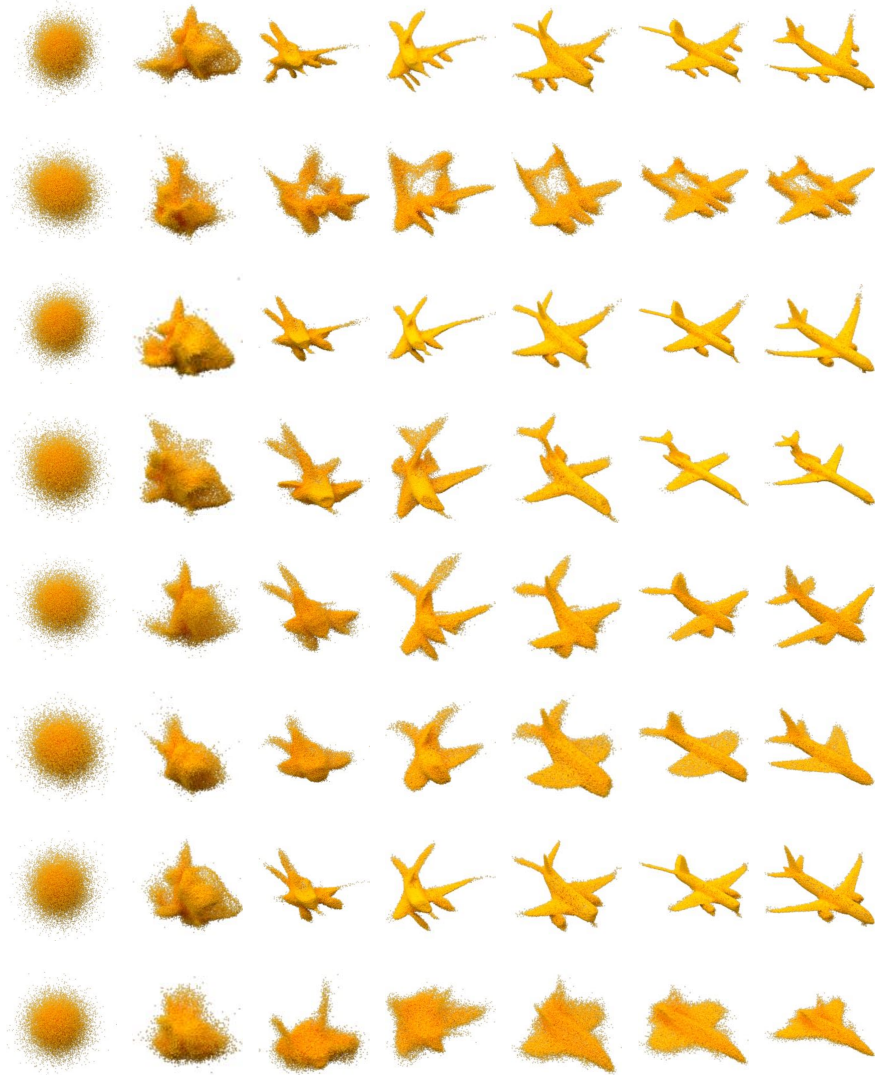}
    \caption{Generating flow for sampled point clouds from the Airplane model.}
    \label{fig:flow_air}
\end{figure}

In figures \ref{fig:flow_air}---\ref{fig:flow_cha} we provide examples of the generating flow for sampled point clouds from the classes Airplane, Car, and Chair. 
We sample a shape from data, obtain initial Gaussian from the corresponding latent variable, sample $32,768$ points by the flow, and then visualize the evolution of the point cloud across the initial Gaussian and layers 32, 48, 56, 60, 62, 63 of the generative flow network.

\clearpage
\begin{table}[ht!]
  \begin{center}
    \caption{Generative modeling evaluation. JSD and MMD-EMD are multiplied by $10^2$, MMD-CD by $10^4$. 
    Cases where the oracle does not obtain best results are underlined.}
    \label{tab:gen_ext}
    \rotatebox{-90}{
      \resizebox{1.30\textwidth}{!}{
\setlength{\tabcolsep}{4pt}
\begin{tabular}{|l|l|c|c|c|c|c|c|c|}
\hline
    &   & JSD$\downarrow$ & \multicolumn{2}{c|}{MMD$\downarrow$} & \multicolumn{2}{c|}{COV$\uparrow$, \%} & \multicolumn{2}{c|}{1-NNA$\downarrow$, \%} \\
Category & Model &                                        & CD               & EMD             & CD         & EMD        & CD         & EMD        \\
\hline
         & l-GAN-CD \cite{achlioptas18icml}  & 2.76       $\pm$ 0.16 & {\bf 5.69}       $\pm$ 0.04 & 5.16            $\pm$ 0.02 & 39.5       $\pm$ 0.8 & 17.1       $\pm$ 0.6 & 72.9       $\pm$ 0.8 & 92.1       $\pm$ 0.6 \\
         & l-GAN-EMD \cite{achlioptas18icml} & 1.77       $\pm$ 0.13 & 6.05             $\pm$ 0.04 & {\bf 4.15}      $\pm$ 0.02 & 39.7       $\pm$ 1.4 & 40.4       $\pm$ 1.2 & 75.7       $\pm$ 0.6 & 73.0       $\pm$ 1.2 \\
Airplane & PointFlow \cite{yang19iccv}       & 1.42       $\pm$ 0.12 & 6.05             $\pm$ 0.05 & 4.32            $\pm$ 0.01 & {\bf 44.7} $\pm$ 1.2 & {\bf 48.4} $\pm$ 1.0 & {\bf 70.9} $\pm$ 1.0 & {\bf 68.4} $\pm$ 1.0 \\
         & DPF-Nets (Ours)                   & {\bf 0.94} $\pm$ 0.11 & 6.07             $\pm$ 0.04 & 4.26            $\pm$ 0.02 & {\bf 46.8} $\pm$ 1.2 & {\bf 48.4} $\pm$ 0.9 & {\bf 70.6} $\pm$ 1.0 & {\bf 67.0} $\pm$ 1.2 \\
         & Oracle                            & 0.50       $\pm$ 0.04 & \underline{5.97} $\pm$ 0.09 & 3.98            $\pm$ 0.01 & 51.4       $\pm$ 1.0 & 52.7       $\pm$ 1.3 & 49.8       $\pm$ 1.3 & 48.2       $\pm$ 1.1 \\
\hline
         & l-GAN-CD \cite{achlioptas18icml}  & 2.65       $\pm$ 0.07 & {\bf 8.83}       $\pm$ 0.06 & 5.36            $\pm$ 0.01 & 41.3       $\pm$ 0.8 & 15.9       $\pm$ 1.3 & {\bf 62.6} $\pm$ 0.6 & 92.7       $\pm$ 0.4 \\
         & l-GAN-EMD \cite{achlioptas18icml} & 1.31       $\pm$ 0.10 & {\bf 9.00}       $\pm$ 0.08 & {\bf 4.40}      $\pm$ 0.01 & 38.3       $\pm$ 1.2 & 32.9       $\pm$ 0.7 & 65.2       $\pm$ 0.4 & {\bf 63.2} $\pm$ 1.0 \\
Car      & PointFlow \cite{yang19iccv}       & 0.59       $\pm$ 0.02 & 9.53             $\pm$ 0.06 & 4.71            $\pm$ 0.01 & {\bf 42.3} $\pm$ 1.0 & 35.8       $\pm$ 1.3 & 70.1       $\pm$ 0.9 & 74.2       $\pm$ 0.6 \\
         & DPF-Nets (Ours)                   & {\bf 0.45} $\pm$ 0.02 & 9.59             $\pm$ 0.04 & 4.61            $\pm$ 0.01 & {\bf 43.4} $\pm$ 0.9 & {\bf 45.8} $\pm$ 1.0 & 70.3       $\pm$ 0.6 & {\bf 64.3} $\pm$ 1.5 \\
         & Oracle                            & 0.37       $\pm$ 0.03 & \underline{9.24} $\pm$ 0.06 &\underline{4.56} $\pm$ 0.01 & 52.8       $\pm$ 1.1 & 52.7       $\pm$ 0.9 & 50.9       $\pm$ 1.1 & 50.5       $\pm$ 1.2 \\
\hline
         & l-GAN-CD \cite{achlioptas18icml}  & 3.65       $\pm$ 0.09 & {\bf 16.66}      $\pm$ 0.08 & 7.91            $\pm$ 0.02 & 42.3       $\pm$ 0.5 & 17.1       $\pm$ 0.5 & 68.5       $\pm$ 0.5 & 96.5       $\pm$ 0.1 \\
         & l-GAN-EMD \cite{achlioptas18icml} & 1.27       $\pm$ 0.06 & {\bf 16.78}      $\pm$ 0.07 & {\bf 5.75}      $\pm$ 0.01 & 44.3       $\pm$ 0.9 & 43.8       $\pm$ 1.0 & 66.6       $\pm$ 0.6 & 67.8       $\pm$ 0.7 \\
Chair    & PointFlow \cite{yang19iccv}       & 1.51       $\pm$ 0.11 & 17.15            $\pm$ 0.10 & 6.20            $\pm$ 0.01 & 43.3       $\pm$ 0.8 & {\bf 46.5} $\pm$ 1.0 & 67.0       $\pm$ 0.3 & 70.4       $\pm$ 0.6 \\
         & DPF-Nets (Ours)                   & {\bf 1.01} $\pm$ 0.06 & 17.08            $\pm$ 0.11 & 6.14            $\pm$ 0.01 & {\bf 46.9} $\pm$ 0.8 & {\bf 48.5} $\pm$ 1.1 & {\bf 63.5} $\pm$ 1.3 & {\bf 64.8} $\pm$ 0.7 \\
         & Oracle                            & 0.49       $\pm$ 0.01 & 16.39            $\pm$ 0.07 & 5.71            $\pm$ 0.01 & 52.8       $\pm$ 0.8 & 53.4       $\pm$ 1.1 & 49.7       $\pm$ 0.7 & 49.6       $\pm$ 0.9 \\
\hline
\end{tabular}
}
    }
  \end{center}
\end{table}

\clearpage
\begin{figure}[h!]
    \centering
    \input{sparse_autoenc.tex}
    \caption{Qualitative comparison in the autoencoding task with sparse inputs. Left to right: reconstructions from l-GAN-CD, l-GAN-EMD, AtlasNet, PointFlow, DPF-Nets (orig.), DPF-Nets (norm.), and ground-truth.}
    \label{fig:sparse_ae_samples}
\end{figure}

\clearpage
\begin{figure}[h!]
    \centering
    \input{dense_autoenc.tex}
    \caption{Qualitative comparison in the autoencoding task with dense inputs. Left to right: reconstructions from l-GAN-CD, l-GAN-EMD, AtlasNet, PointFlow, DPF-Nets (orig.), DPF-Nets (norm.), and ground-truth.}
    \label{fig:dense_ae_samples}
\end{figure}

\clearpage
\begin{figure}[h!]
    \centering
    \input{airplane_samples.tex}
    \caption{Random samples from  models trained on the Airplane class. 
    Columns 1---3  samples from PointFlow, columns 4---6  samples from DPF-Net.}
    \label{fig:samples_air}
\end{figure}

\clearpage
\begin{figure}[h!]
    \centering
    \input{car_samples.tex}
    \caption{Random samples from  models trained on the Car class. 
    Columns 1---3  samples from PointFlow, columns 4---6  samples from DPF-Net.}
    \label{fig:samples_car}
\end{figure}

\clearpage
\begin{figure}[h!]
    \centering
    \input{chair_samples.tex}
    \caption{Random samples from  models trained on the Chair class. 
    Columns 1---3  samples from PointFlow, columns 4---6  samples from DPF-Net.}
    \label{fig:samples_cha}
\end{figure}

\clearpage

\clearpage
\begin{figure}[h!]
    \centering
    \input{car_interpolations.tex}
    \caption{Interpolations between Car data samples.}
    \label{fig:int_car}
\end{figure}

\clearpage
\begin{figure}[h!]
    \centering
    \input{chair_interpolations.tex}
    \caption{Interpolations between Chairs data samples.}
    \label{fig:int_cha}
\end{figure}

\clearpage

\clearpage
\begin{figure}[h!]
    \centering
    \input{car_flows.tex}
    \caption{Generating flow for sampled point clouds from the Car model.}
    \label{fig:flow_car}
\end{figure}

\clearpage
\begin{figure}[h!]
    \centering
    \input{chair_flows.tex}
    \caption{Generating flow for sampled point clouds from the Chair model.}
    \label{fig:flow_cha}
\end{figure}


\end{document}